# A Model for Forecasting Air Quality Index in Port Harcourt Nigeria Using Bi-LSTM Algorithm

O. E. Taylor[1] & P. S. Ezekiel[2]

*taylor.onate@ust.edu.ng, Tel.: +2348034448978; ezekielpromise27@gmai.com Tel.: +2348109809293*

[1,2]Department of Computer Science, Rivers State University, Port Harcourt, Nigeria

**Abstract –** The release of toxic gases by industries, emissions from vehicles, and an increase in the concentration of harmful gases and particulate matter in the atmosphere are all contributing factors to the deterioration of the quality of the air. Factors such as industries, urbanization, population growth, and the increased use of vehicles contribute to the rapid increase in pollution levels, which can adversely impact human health. This paper presents a model for forecasting the air quality index in Nigeria using the Bi-directional LSTM model. The air pollution data was downloaded from an online database (UCL). The dataset was pre-processed using both pandas tools in python. The pre-processed result was used as input features in training a Bi-LSTM model in making future forecasts of the values of the particulate matter Pm2.5, and Pm10. The Bi-LSTM model was evaluated using some evaluation parameters such as mean square error, mean absolute error, absolute mean square, and $R^2$ square. The result of the Bi-LSTM shows a mean square error of 52.99%, relative mean square error of 7.28%, mean absolute error of 3.4%, and $R^2$ square of 97%. The model. This shows that the model follows a seamless trend in forecasting the air quality in Port Harcourt, Nigeria.

## 1.0 Introduction

The oil-rich nation The Niger Delta is situated on the delta of the "River Niger" in southern Nigeria. The Niger Delta consists of nine oil-producing states, including Abia, Akwa-Ibom, Bayelsa, Cross River, Delta, Edo, Imo, Ondo, and Rivers, whose lands and rivers provide all of Nigeria's crude oil. Its primary biological zones consist of coastal Barrier islands, freshwater swamp forests, lowland rain forests, and mangrove forests. The Niger Delta encompasses 70,000 square kilometres of southern Nigeria and is the flood plain through which the rivers Benue and Niger drain into the Atlantic Ocean. Essentially a network of creeks linking rivers, rivulets, and streams, including the rivers Benin, Bonny, Brass, Cross, and Nun, it acts as a repository for economically valuable national resources.

In recent years, rapid population growth, urbanization, and changes in people's lifestyles have significantly contributed to an increase in urban areas' levels of air pollution. The public's attention has recently been focused on the regulation of environmental pollution. It is a widely held belief that urban air pollution has a direct impact on human health, particularly in developing countries as well as industrialized nations where measures to improve air quality are either not available or are only minimally implemented or enforced [1]. Recent studies have shown substantial evidence that exposure to atmospheric pollutants has strong links to air pollution and human health, there will invariably be an increase in the costs associated with receiving medical care, particularly in terms of admissions to hospitals and trips to emergency rooms. This will have a negative effect on the economy [2].

The release of toxic gases by industries, emissions from vehicles, and an increase in the concentration of harmful gases and particulate matter in the atmosphere are all contributing factors to the deterioration of the quality of the air. Factors such as industries, urbanization, population growth, and the increased use of vehicles are contributing to the rapid increase in pollution levels, which can have an adverse impact on human health. The most important contributor to the rising levels of air pollution is a particulate matter [3]. Due to this, there is a requirement for the measurement and analysis of real-time monitoring of the quality of the air in order for appropriate decisions to be made in a timely manner. Air pollution is a contributing factor to the hazardous condition that has a deleterious effect on living things. It is one of the major concerns shared by people all over the world.

Air pollution is a concern shared by international organizations, national governments, and the mainstream media all over the world. Plants, air, and water can become tainted with harmful substances if natural resources are exploited at a rate that is greater than nature's capacity to regenerate them on its own. In addition to the activities of humans, there are a few naturally occurring cycles that, when they occur, also result in the release of potentially harmful substances [4].

The critical situation involving the world's air pollution is a major problem that endangers our planet. It can have a variety of negative effects on people's health as well as the living ecosystem. In point of fact, research conducted by the State of Global Air (SOGA) has demonstrated that being subjected to high levels of air pollution lowers a person's life expectancy by an average of 20 months across the globe, and by 18 months in North Africa and the Middle East. The most common types of pollutants in the air are gaseous pollutants (such as carbon dioxide ($CO_2$), carbon monoxide (CO), sulphur dioxide ($SO_2$), ozone ($O_3$), nitrogen oxide (NO), and nitrogen dioxide ($NO_2$)), and particulate matters (PM), which are a complex mixture of solid and liquid droplets (e.g., PM2.5, PM10). Particulate matter is the culprit behind a wide variety of respiratory illnesses, including asthma, chronic obstructive pulmonary disease (COPD), and respiratory infections [5].

## 2. Related Works

[6] utilized Recurrent Neural Network (RNN) and Support Vector Classifier (SVC) in predicting the quality of air index. The stated algorithms were trained using a Central Pollution Control Board (CPCB) dataset that was collected from the ministry of environment. The proposed models archived accuracy results of 91.62% and 97.3%.

[2] made use of three machine learning algorithms in monitoring the level of air pollution. The algorithms M5P models, Support Vector Classifier (SVC), and Artificial Neural Network. Their

experimental result shows that using different features in multivariate modelling with the M5P algorithm yields the best forecasting performances at 31.4%

[7] utilized various classification and regression approaches such as Linear SDG Regression, Random, Decision Tree Regression, Forest Regression, Regression, Support Vector Regression, Artificial Neural Networks, Adaptive Boosting Regression, and Gradient Boosting Regression to forecast the Air Quality Index of major pollutants like PM2.5, PM10, CO, NO2, SO2 and O3. Mean square error, mean absolute error and R2, were used in evaluating the performance of the models. The result shows that Support Vector Regression and Artificial Neural Networks had the best outcome in predicting the quality of air in Delhi city.

[8] proposed an IoT-based system for monitoring the quality of air in a city. The authors start by acquiring an IoT device called Raspberry Pi. They connected the device with some external sensors, which they used in capturing data for both bad and good air. The collected data was then used in training a machine learning model. The performance of the model was measured using Mean Square Error and Mean Absolute Error. The model achieved a Mean Square Error of 0.84% and a Mean Absolute Error of 3.97%.

[9] proposed a set of sensor arrays based on the Internet of Things to detect eight different pollutants, including NH3, CO, NO2, CH4, CO2, PM2.5, temperature, and humidity. To begin the analysis process, the sensor array takes readings of the pollutant levels and sends them to the cloud server through a series of gateways. The information that was gathered was used in training an Artificial Algae Algorithm (AAA) and an Elman Neural Network (ENN) model, both of which were used to make forecasts about the quality of the air at future time stamps. In order to determine the ENN model's parameter values in the most accurate way possible, the parameter tuning technique known as the AAA is utilized. The results of the experiments show that the proposed ETAPM-AIT model has superior performance to the more recent techniques.

[10] developed an air pollution monitoring system using Arduino Board. The Arduino board was used in collecting various sensors/ signals. For real-time testing, the collected data was used as input for a machine-learning model. Their experiment shows that their proposed model works efficiently.

Air quality modeling methods must be used to study how industrial operations affect the dispersion of air pollutants and how those pollutants affect the surroundings that receive them. The air quality at a large cement plant complex in Ibese, Ogun State, Nigeria, was studied, and dispersion modeling was performed. It was determined that the manufacturing facility's primary point sources of atmospheric emissions were the process facilities and the utility air emission control equipment. Along the plant's perimeter, gaseous pollutants from all the detected point emission sources in the plant were 0.01-276.13% above statutory limits; 24-hour PM10 was 14.32% above the limit and yearly PM10 was 4.90% above the limit. All fence line locations register 1-hour SO2 levels below the limit, despite all point sources operating at full capacity. An ambient air quality monitoring system and air emission dispersion modeling were used to track the air quality [11].

This research looked for links between pollutant geographical variation and dispersion and the presence of human-caused industries and other anthropogenic activities in the region. Areas in

Egbema, Orlu, Okigwe, and Orlu in Imo State, Nigeria, form part of the research area. The Gasman Air Monitors were utilised to detect levels of sulphur dioxide (SO2), nitrogen dioxide (NO2), and carbon monoxide (CO), while the Haze-dust Particulate Monitor 10 m was employed to collect data on particulate matter (PM10) three times daily, in the morning, afternoon, and evening. Similarly, H2S was measured using an Aeroqual Gas Monitor Model 300, and volatile organic compound concentrations (VOCs) were measured using an Ibrid MX6 Gas Monitor [12].

[13] evaluated the standard of air in key sawmill locations in Ilorin, Kwara State, Nigeria. The levels of PM10 and formaldehyde in the air were measured using a BLATN BR—Smart Series air quality monitor, a BLATN BR—Smart Series air quality multimeter measuring PM2.5, VOCs, temperature, and humidity, and a hand-held mobile multi-gas monitor, model AS8900, measuring combustible levels and oxygen levels. This study's findings demonstrate that average levels of carbon monoxide (CO), oxygen (O2), and other measured parameters like formaldehyde (HcHo), etc., are typically below and within the permitted range of National and International regulatory guidelines for air quality indices. The average amounts of volatile organic compounds (VOCs), particulate matter (PM2.5), particulate matter (PM10), and combustible (LEL) are all above national and international guidelines, however, there are outliers.

## 3. Design Methodology

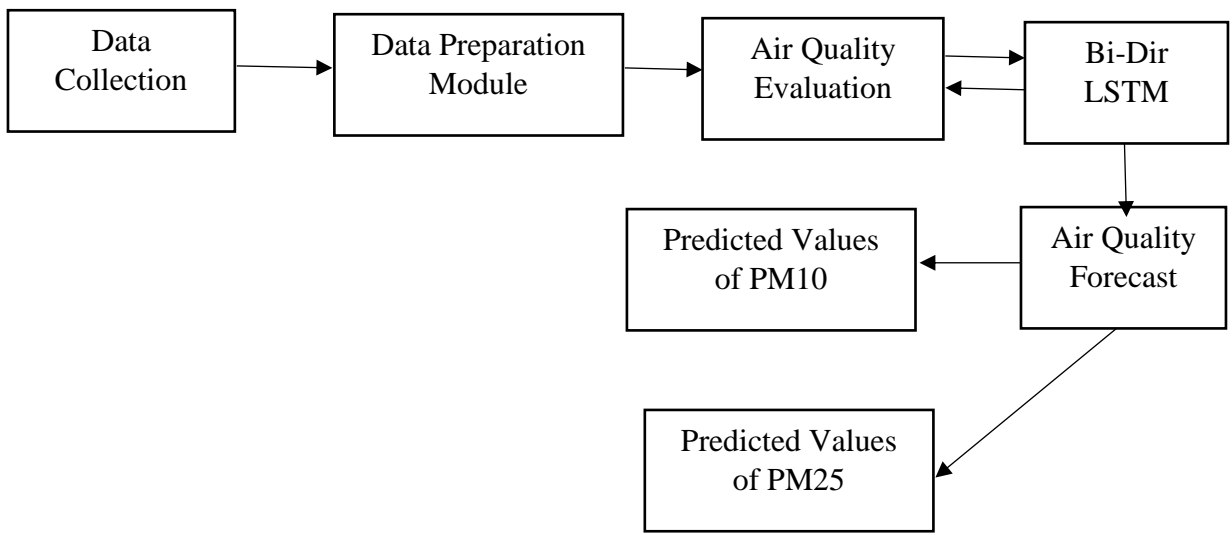

Figure 3.2. Architectural Design of the Proposed System

**Data Collection:** The data used here was downloaded from an online database. The data was collected using portable digital handheld particle monitors. This was used to measure the levels of PM2.5 and PM10 in Nigeria. The equipment has a laser optical sensor that can detect and measure particle concentrations up to 1 gm-3. It has already been calibrated to make sure the quality is good. Particles were collected during both the rainy and dry months of 2018 (May–December). At a height of 1.5 m above ground, the sampler was retained within the safe range for human respiration. The screen of the gadget shows the collected PM concentration readings. Sampling was conducted 24 hours a day, 7 days a week for 8 months. Meteorological factors and data were used together to figure out the AQI, particle pollution episodes, the effect of weather on PM concentration, and the relationship between PM and meteorological factors.

Table 1: Dataset Sample

| S/no. | date | wd | ws | temp | rh | rfall | pm25 | pm10 |
|---|---|---|---|---|---|---|---|---|
| 0 | 2018-10-05 18:00:00 | 180.0 | 0.3 | 24.5 | 94.0 | 0.0 | 5 | 22 |
| 1 | 2018-10-05 22:00:00 | 315.0 | 1.7 | 24.9 | 95.0 | 0.0 | 26 | 100 |
| 2 | 2018-10-05 23:00:00 | 45.0 | 1.0 | 24.4 | 96.0 | 0.0 | 36 | 138 |
| 3 | 2018-11-05 00:00:00 | 315.0 | 0.7 | 24.3 | 96.0 | 0.0 | 21 | 80 |
| 4 | 2018-11-05 01:00:00 | 45.0 | 0.3 | 24.3 | 96.0 | 0.0 | 25 | 96 |
| ... | ... | ... | ... | ... | ... | ... | ... | ... |
| 4805 | 2018-12-31 19:00:00 | 225.0 | 2.4 | 29.7 | 59.0 | 0.0 | 55 | 145 |
| 4806 | 2018-12-31 20:00:00 | 180.0 | 1.4 | 29.1 | 61.0 | 0.0 | 47 | 129 |
| 4807 | 2018-12-31 21:00:00 | 225.0 | 1.0 | 28.6 | 63.0 | 0.0 | 64 | 167 |
| 4808 | 2018-12-31 22:00:00 | 205.0 | 1.4 | 28.2 | 65.0 | 0.0 | 64 | 165 |
| 4809 | 2018-12-31 23:00:00 | 225.0 | 1.0 | 28.0 | 67.0 | 0.0 | 75 | 197 |

**Data Preparation Module:** In this module, the cleaning and the pre-processing of the data is carried out. Here, the data pre-processing that will be carried out has to do with the checking and removal of Nan values, and data scaling. Data scaling is an important pre-processing step when working with both machine learning and. Data scaling can be achieved by normalizing or standardizing real-valued input and output variables.

**Air Quality Evaluation:** This module is used in evaluating the quality of the air, if it is good, moderate, unhealthy, or hazardous. For instance, an AQI of 100 indicates that the concentration of pollutants in the air is above the standard, making it unhealthy for some groups of people and eventually everyone. In this investigation, we used PM2.5 and PM10 concentration data to determine the AQI. Using the EPA's standard formula presented in the equation, an AQI was calculated (3.2). The thresholds for PM2.5 and PM10 AQI are listed in Table 2.

**Bi-Directional LSTM Models:** We made use of the bi-directional LSTM algorithm in making our proposed method not just have the sequence of flow of information in one direction (backpropagation) but have the sequence of flow of information in both directions (forward and backward propagation). The process of calculating the current state of our proposed method can be seen in Eqn. (1) and applying the activation function to the current state can be seen in Eqn. (2). The output of the bi-directional model can be seen in Eqn. (3).

$$h_t = f(h_{t-1}, x_t) \quad \ldots \quad \text{Eqn. (1)}$$

$$h_t = \tanh(W_{hh} h_{t-1} + W_{xh} \quad \ldots \quad \text{Eqn. (2)}$$

$$\hat{y}^{<t>} = g(W_y [\vec{a}^{<t>}, \overleftarrow{a}^{<t>}] + b_y) \ldots \quad \text{Eqn. (3)}$$

From the above equations, the following are the definitions of the notations:

$h_t$ represents the current state of the model

$h_{t-1}$ represents the previous state of the model

$x_t$ represents the input features of the air pollution data.

tanh represents the activation function

$W_{hh}$ represents the weight at the recurrent neuron

$W_{xh}$ represents weight at input neuron

$\hat{y}^{<t>}$ represents the output (which is the values of the predicted particulate matters)

$W_y$ represents the weight of the output

$\vec{a}^{<t>}$ represents the activation function at the forward direction

$\overleftarrow{a}^{<t>}$ represents the activation function of the backward direction

$b_y$ represents the base of the output

## 4. Experimental Result

An experiment was carried out on Jupyter Notebook Technology. The experimental phase is made up of two phases. The phases are:

**4.1 Exploratory Data Analysis:** This has to do with the reading of the air pollution data into the Jupyter Notebook environment, checking the relationships between features of the dataset, checking the importance of each features of the particulate matters, and metrological parameters towards the effect of air pollution on the environment, and also the effect of the particulate matters through charts and graphs. We checked the relationship between each feature of the air pollution data by plotting a correlation matrix in python. This was achieved using the corr() in python. The correlation of the dataset features can be seen in Table 2. After we checked for the relationship between the dataset features, we then checked the importance of each features, and their effect on air particulate matter. This was achieved using Adfuller in python. This can be seen in Figure 2. Further analysis was also carried out on the dataset, such as what degree of relative humidity has more effect on air particulate matters, and what direction of the wind has more effects on air particulate matters. This can be found in Figure 3 and Figure 4. Figure 5 and Figure 6 shows the increase in the air particulate matter over time. Figure 7 and Figure 8 also shows the daily (Day and Night) effect of the particulate matter Pm10 and Pm2.5.

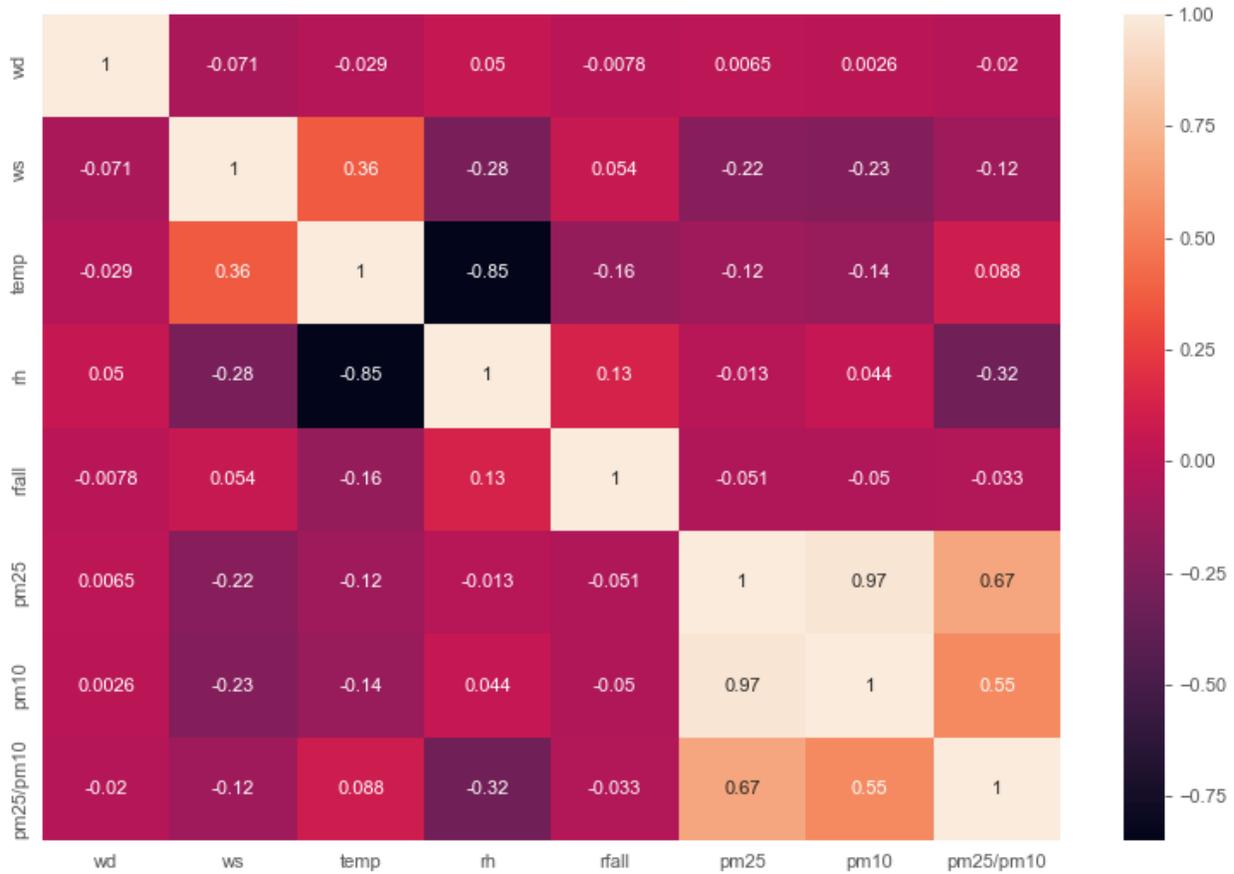

Figure 1: Correlation Matrix of the Dataset

Table 2: Important Features of the Dataset

```
============== date ===============
ADF Statistic: 0.090720
p-value: 0.965422
non-stationary
============== wd ===============
ADF Statistic: -26.998923
p-value: 0.000000
Stationary
============== ws ===============
ADF Statistic: -9.726474
p-value: 0.000000
Stationary
============== temp ===============
ADF Statistic: -6.197940
p-value: 0.000000
Stationary
============== rh ===============
ADF Statistic: -4.139683
p-value: 0.000832
Stationary
============== rfall ===============
ADF Statistic: -51.857198
p-value: 0.000000
Stationary
============== pm25 ===============
ADF Statistic: -6.521411
p-value: 0.000000
Stationary
============== pm10 ===============
ADF Statistic: -7.078704
p-value: 0.000000
Stationary
============== pm25/pm10 ===============
ADF Statistic: -3.604770
p-value: 0.005670
Stationary
```

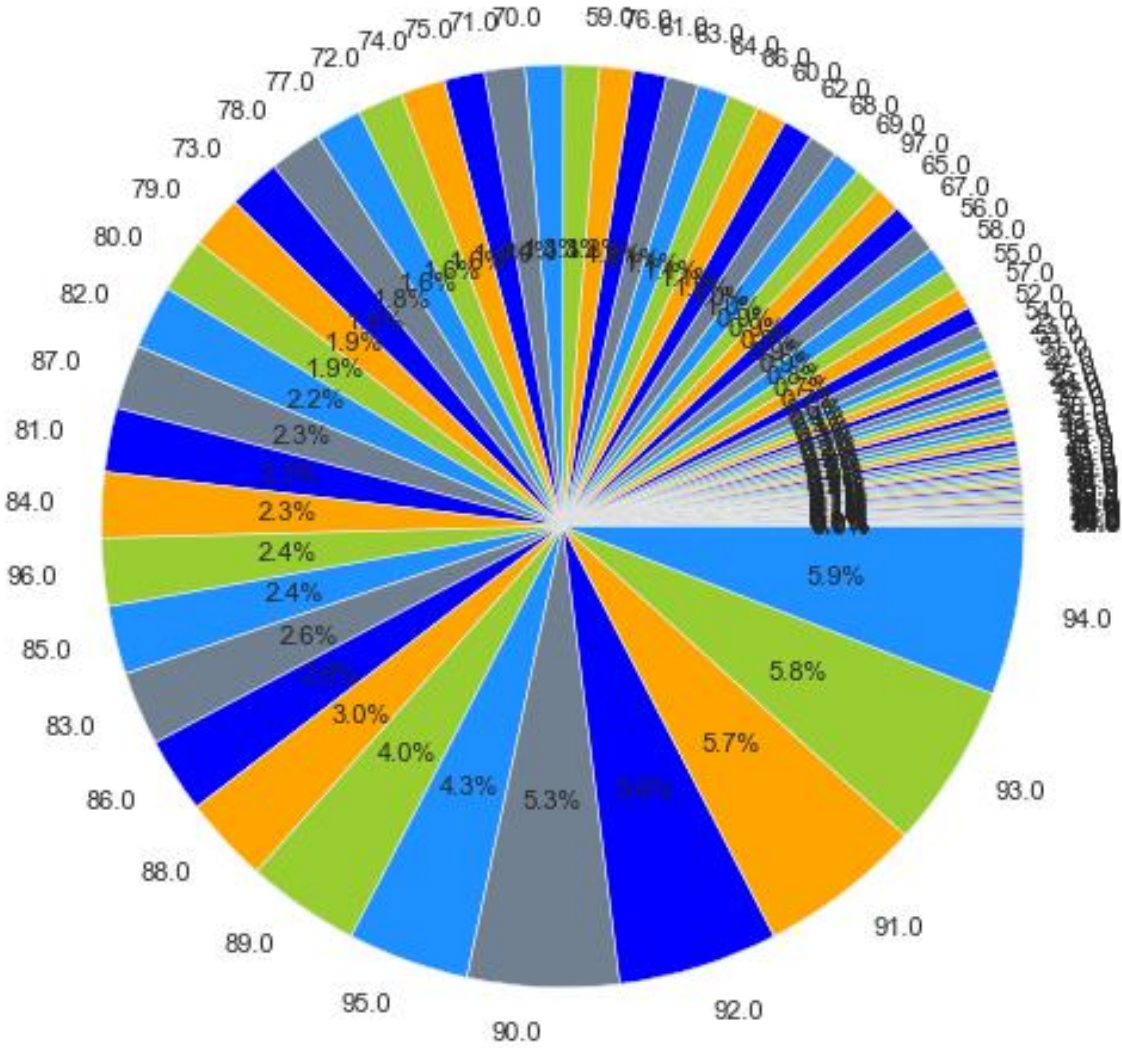

Figure 3: Effect of the Degree of Relative Humidity towards Air Pollution

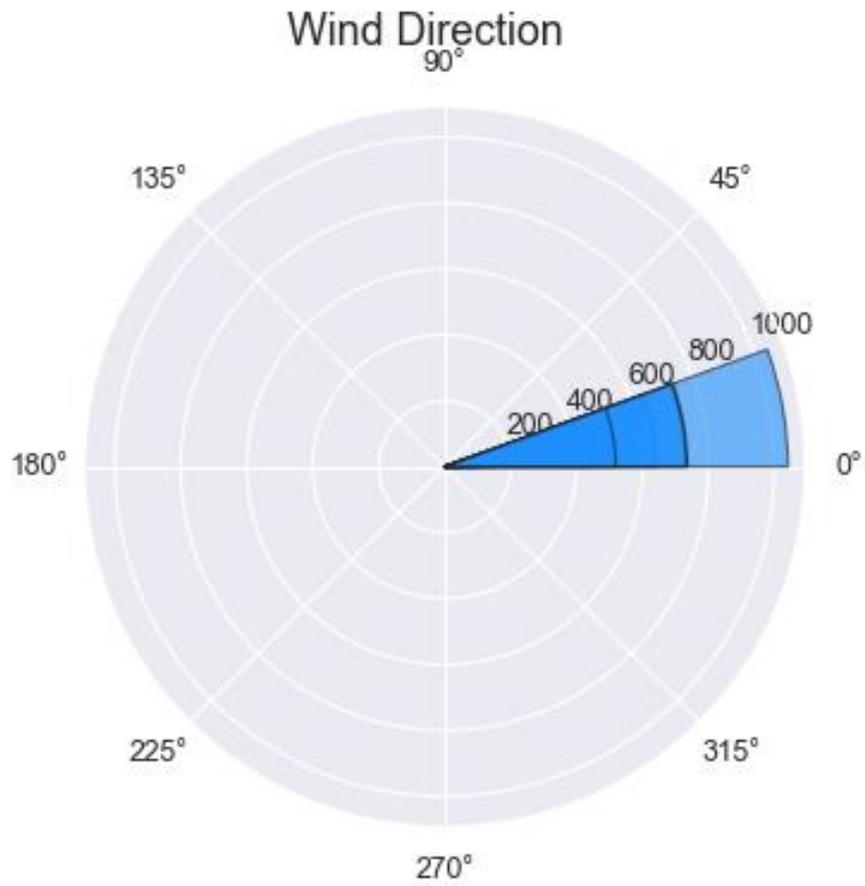

Figure 4: Wind direction towards the effect of air particulate matters

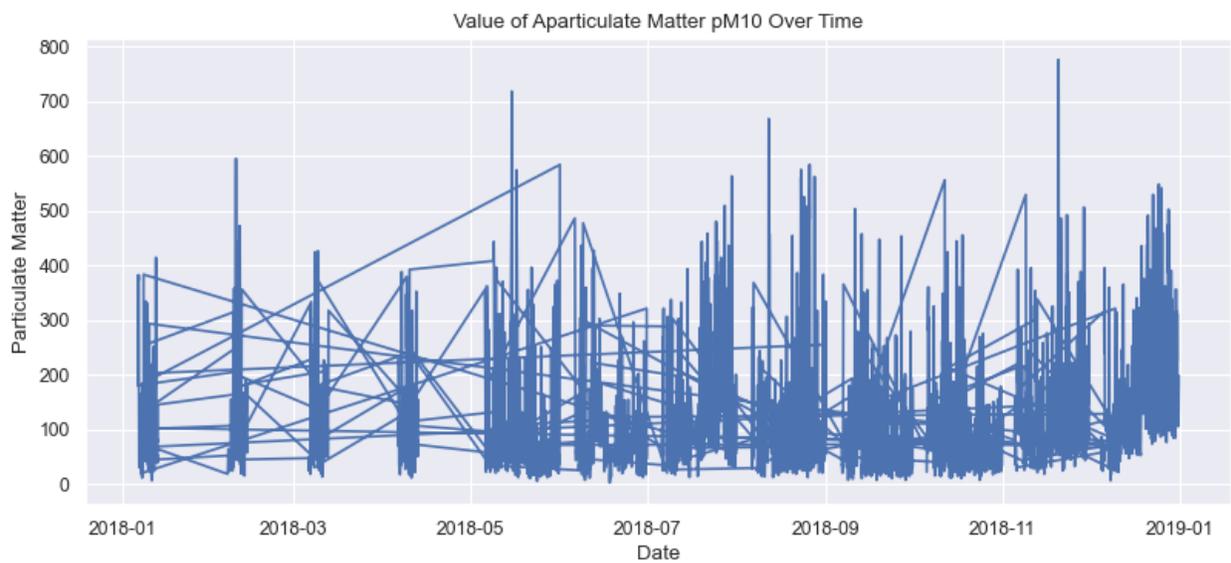

Figure 5: Value of particulate matter Pm10 over time

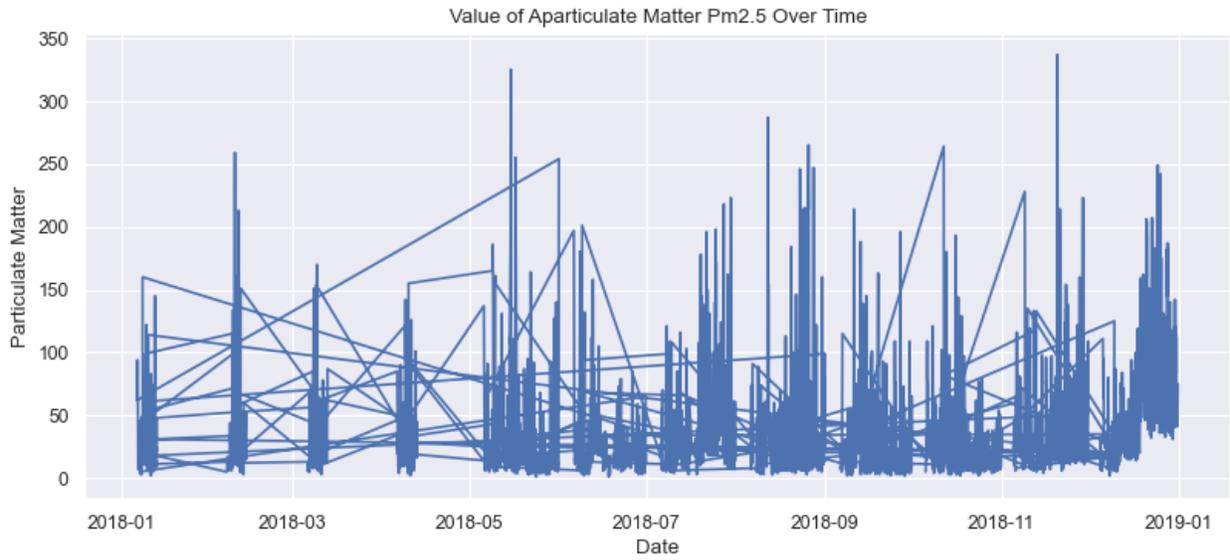

Figure 6: Value of particulate matter Pm2.5 over time

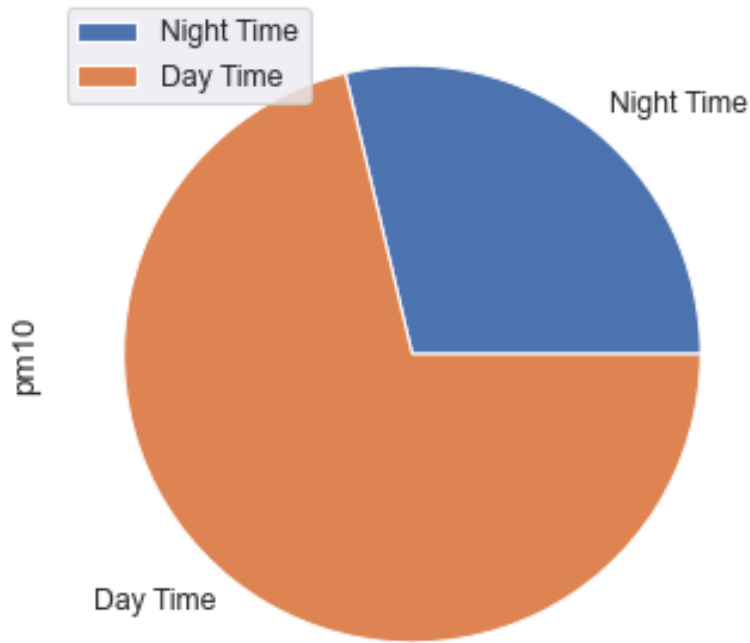

Figure 7: Increase in the air particulate matter Pm10 over time.

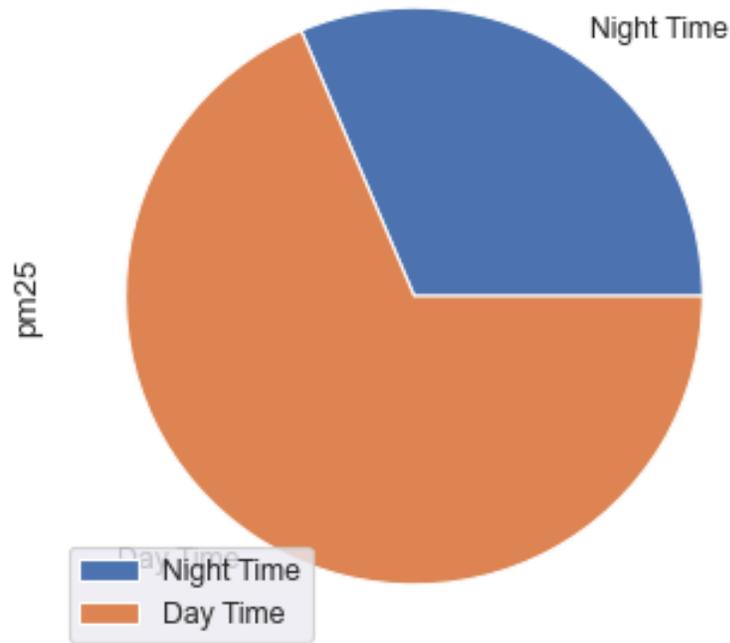

Figure 8: Increase in the air particulate matter Pm2.5 over time.

### 4.2 Training of the Regression Model

The model was trained using Bi Directional Long-Short Term Memory. The Bi-LSTM model was trained using the four layers. The first layer contains an input neuron of 20 and used relu as activation function. The second layer contains an input neuro of 10, and activation function of tanh. The third layer contain an input neuron of 1024, and an activation function of relu, and finally the fourth layer being the output layer used sigmoid as activation function. Other hyper parameters used in training the model are loss= mean_squared_error, optimizer=adma, epoch, 20 and batch_size=32. The training result which displays the mean squared error for both training and validation test. This can be seen in Table 3 After training, the model was used in making prediction on the test data, the predicted result on the test data can be seen Figure 9.

Table 3. Training process of the Bi-Directional LSTM model

| |
|---|
| Epoch 1/20 |
| 35036/35036 [==============================] - 30s 871us/step - loss: 0.0021 |
| Epoch 2/20 |
| 35036/35036 [==============================] - 29s 815us/step - loss: 0.0011 |
| Epoch 3/20 |
| 35036/35036 [==============================] - 28s 788us/step - loss: 9.0931e-04 |
| Epoch 4/20 |
| 35036/35036 [==============================] - 27s 778us/step - loss: 8.4520e-04 |
| Epoch 5/20 |
| 35036/35036 [==============================] - 27s 783us/step - loss: 8.4552e-04 |
| Epoch 6/20 |
| 35036/35036 [==============================] - 28s 788us/step - loss: 8.4213e-04 |
| Epoch 7/20 |
| 35036/35036 [==============================] - 28s 795us/step - loss: 8.2123e-04 |
| Epoch 8/20 |
| 35036/35036 [==============================] - 27s 776us/step - loss: 8.2782e-04 |
| Epoch 9/20 |
| 35036/35036 [==============================] - 27s 773us/step - loss: 8.1744e-04 |
| Epoch 10/20 |
| 35036/35036 [==============================] - 27s 769us/step - loss: 7.9839e-04 |
| Epoch 11/20 |
| 35036/35036 [==============================] - 27s 761us/step - loss: 8.0556e-04 |
| Epoch 12/20 |
| 35036/35036 [==============================] - 27s 769us/step - loss: 8.0480e-04 |
| Epoch 13/20 |
| 35036/35036 [==============================] - 27s 770us/step - loss: 8.0340e-04 |
| Epoch 14/20 |
| 35036/35036 [==============================] - 27s 769us/step - loss: 8.0428e-04 |
| Epoch 15/20 |

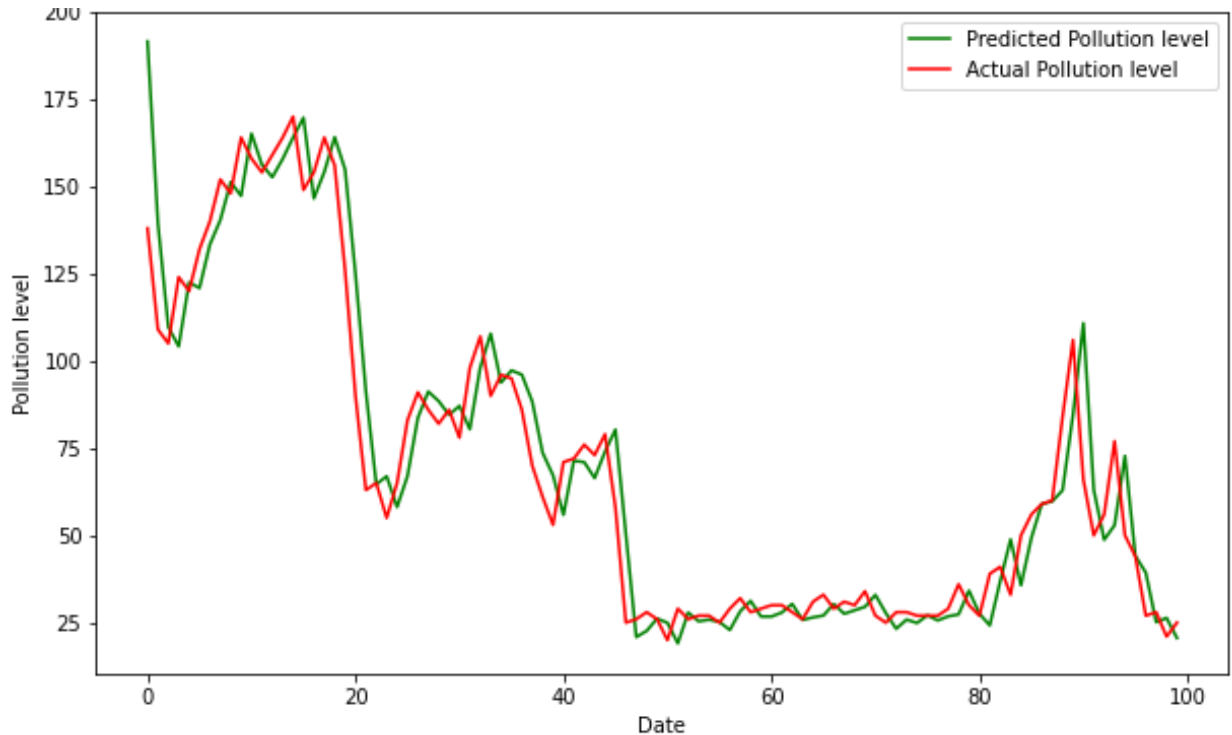

Figure 9: Actual Result of Particulate Matter Pm25 and Pm10

## 5. Discussion of Results

From the experiment conducted, Figure 1 shows the correlation matrix between the features of the air pollution dataset. The correlation matrix is used in checking features of the air pollution dataset that are most correlated (Features that are have relationship between each other). Table 2 shows the statistical analysis of the variables (columns) on the dataset. The p-value determines how significant the variables in the columns are with respect to the target variable. The p-value less or equal to 0.005 signifies that the variable (column) is significant while the values greater than 0.005 is significantly not important. Figure 3 shows the degree of relative humidity with respect to its effect on particulate matters. The pie chart in Figure 3 shows the level of the relative humidity in degrees that has more effect on particulate matters. Figure 4 shows the direction of the wind in degrees that contributes to the values of the particulate PM10 and PM2.5. Figure 5 shows the increase of the particulate matter PM10 over time, and Figure 6 also shows the increase of the particulate matter PM2.5 over time. Figure 7 shows that the value of the particulate matter PM10 has more effect on day more than night. Figure 8 shows that the value of the particulate matter PM2.5 has more effect on day more than night. Table 3 shows the training of the Bi-directional LSTM model. The table shows the loss and mean absolute error obtained by the model at each training step. The Bi-directional LSTM model was trained on 20 epoch (training steps). Figure 9 shows the actual value of the particulate matters and the predicted values of the air particulate matters. The graph actually shows the changes of the particulate matter PM10, and PM2.5.

## 6. Conclusion and Future Work

This paper presents a model for forecasting the air quality index in Nigeria using the Bi-directional LSTM model. To forecast Nigeria's air quality indexTh, we carried out an experiment on Jupyter Notebook Technology. The air pollution data was downloaded from an online database (UCL). The dataset was pre-processed using both pandas tools in python. The pre-processed result was used as input features in training a Bi-LSTM model in making future forecasts of the values of the particulate matter Pm2.5, and Pm10. The Bi-LSTM model was evaluated using some evaluation parameters such as mean square error, mean absolute error, absolute mean square, and $R^{^2}$ square. The result of the Bi-LSTM shows a mean square error of 52.99%, relative mean square error of 7.28%, mean absolute error of 3.4%, and $R^2$ square of 97%. The model. This shows that the model follows a perfect trend in forecasting the air quality in Port Harcourt Nigeria.

The work can further be improved by making a forecast of the environmental effect of the air particulate matter and the health effect on the residents of Nigeria.